\documentclass[10pt,twocolumn,letterpaper]{article}

\usepackage[pagenumbers]{cvpr} % To force page numbers, e.g. for an arXiv version

\makeatletter
\@namedef{ver@everyshi.sty}{}
\makeatother
\usepackage{tikz}

\usepackage{tikz}
\usepackage{comment}
\usepackage{color}

\usepackage{graphicx}
\usepackage{amsmath}
\usepackage{amssymb}
\usepackage{booktabs}
\usepackage{rotating}
\usepackage{multirow}
\usepackage{amsthm}

\usepackage{algorithm,algorithmic}
\usepackage{subcaption}
\usepackage{bbm}
\usepackage{bm}
\graphicspath{ {images/} }

\DeclareMathOperator*{\argmin}{arg\,min}
\newcommand{\norm}[1]{\left\lVert#1\right\rVert}

\usepackage[pagebackref,breaklinks,colorlinks]{hyperref}

\usepackage[capitalize]{cleveref}
\crefname{section}{Sec.}{Secs.}
\Crefname{section}{Section}{Sections}
\Crefname{table}{Table}{Tables}
\crefname{table}{Tab.}{Tabs.}

\begin{document}

%\title{Fast and Efficient Detection of Adversarial Examples in Deep Hashing based Image Retrieval}

\title{Unsupervised Multi-Criteria Adversarial Detection in Deep Image Retrieval}

\author{Yanru Xiao\\
Old Dominion University\\
Norfolk, VA, USA\\
{\tt\small yxiao002@odu.edu}
\and
Cong Wang\thanks{Correspondence to cwang85@zju.edu.cn.}\\
Zhejiang University\\
Hangzhou, China\\
{\tt\small cwang85@zju.edu.cn}
\and
Xing Gao\\
University of Delaware\\
Newark, DE, USA\\
{\tt\small xgao@udel.edu}
}
\maketitle

%%%%%%%%% ABSTRACT
\begin{abstract}
The vulnerability in the algorithm supply chain of deep learning has imposed new challenges to image retrieval systems in the downstream. Among a variety of techniques, deep hashing is gaining popularity. As it inherits the algorithmic backend from deep learning, a handful of attacks are recently proposed to disrupt normal image retrieval. Unfortunately, the defense strategies in softmax classification are not readily available to be applied in the image retrieval domain. In this paper, we propose an efficient and unsupervised scheme to identify unique adversarial behaviors in the hamming space. In particular, we design three criteria from the perspectives of hamming distance, quantization loss and denoising to defend against both untargeted and targeted attacks, which collectively limit the adversarial space. The extensive experiments on four datasets demonstrate $2-23\%$ improvements of detection rates with minimum computational overhead for real-time image queries.
% \keywords{Deep hashing, adversarial detection, image retrieval.}
\end{abstract}

\vspace{-0.15in}
\section{Introduction}

Powered by neural networks, deep hashing enables image retrieval at a large scale~\cite{dsh,hashnet,dch,zhu2016deep,lin2015deep,csq}. By representing high-dimensional images with compact binary codes, retrieval becomes an efficient similarity computation of Hamming distance. Google~\cite{google}, Bing~\cite{bing}, Pinterest~\cite{pinterest}, Taobao~\cite{taobao} have all incorporated image query as part of their products. Despite of its great success, deep hashing also inherits the vulnerabilities from neural networks~\cite{szegedy2013intriguing} with new attack vectors and effects. By introducing adversarial perturbations either on the query or database images, normal requests can be diverted to an irrelevant (\emph{untargeted attack})~\cite{tao} or specific category (\emph{targeted attack})~\cite{bai2020targeted,wang2021targeted,xiao2021you}, e.g., turning a query of ``husky dog'' into retrieving a branded ``dog food'' so the attacker can advertise their products for free.

\begin{figure}[!t]
    \centering
    \includegraphics[width=0.45\textwidth]{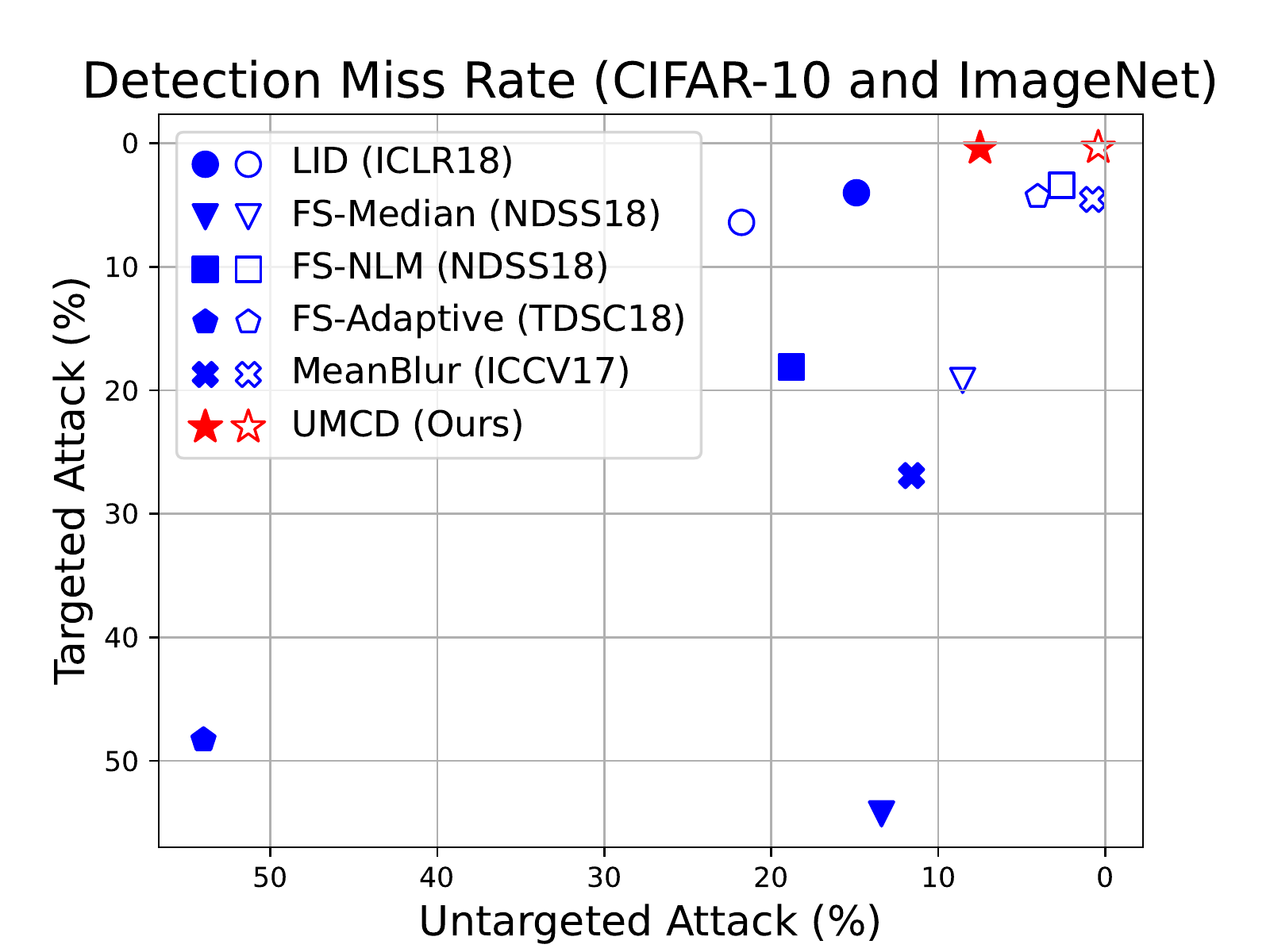}
    \caption{Miss rate of different detections for untageted/targeted adversarial examples in deep hashing. The solid and hollow markers are for CIFAR-10 and ImageNet respectively. The proposed UMCD has the lowest miss rate on both datasets.}
    \label{fig:missrate_compare}
    \vspace*{-0.1in}
\end{figure}

With a handful of efforts on the attack side~\cite{tao,bai2020targeted,wang2021targeted,xiao2021you}, deep hashing still falls short to defend against adversarial examples in the hamming space. \emph{Adversarial training} and \emph{detection} are the two common defenses in softmax classification. Yet, adversarial training has to deal with the non-trivial trade-off between robustness and accuracy~\cite{zhang2019theoretically}. According to our implementation (see appendix), finding the min-max saddle points becomes even more difficult under the hash function, which suffers from a large accuracy loss. On the other hand, detection aims to unveil the adversarial behaviors on different levels of raw pixel~\cite{gong2017adversarial,grosse2017statistical}, feature distribution~\cite{grosse2017statistical,li2017adversarial,ma2018characterizing}, softmax probabilities~\cite{hendrycks2016baseline} and frequency components~\cite{wang2020high} in a \emph{supervised}~\cite{Carrara_2018_ECCV_Workshops} or \emph{unsupervised} manner~\cite{xu2017feature}. Based on the prior knowledge of attack methods, supervised detection trains a classifier to distinguish the adversarial images, but is hard to extrapolate to the unknown attacks. To this end, we pursue the direction of unsupervised anomaly detection in this paper. Different from softmax classification on a closed set of class probabilities, deep hashing maps similar/dissimilar images into binary codes in an open Hamming space. Thus, the focus of our work is to tap into the unique adversarial behaviors in deep hashing to detect both untargeted and targeted attacks.

\begin{figure*}[!t]
\vspace*{-0.09in}
\centering
\hspace*{-0.1in}
\includegraphics[width=1.0\textwidth]{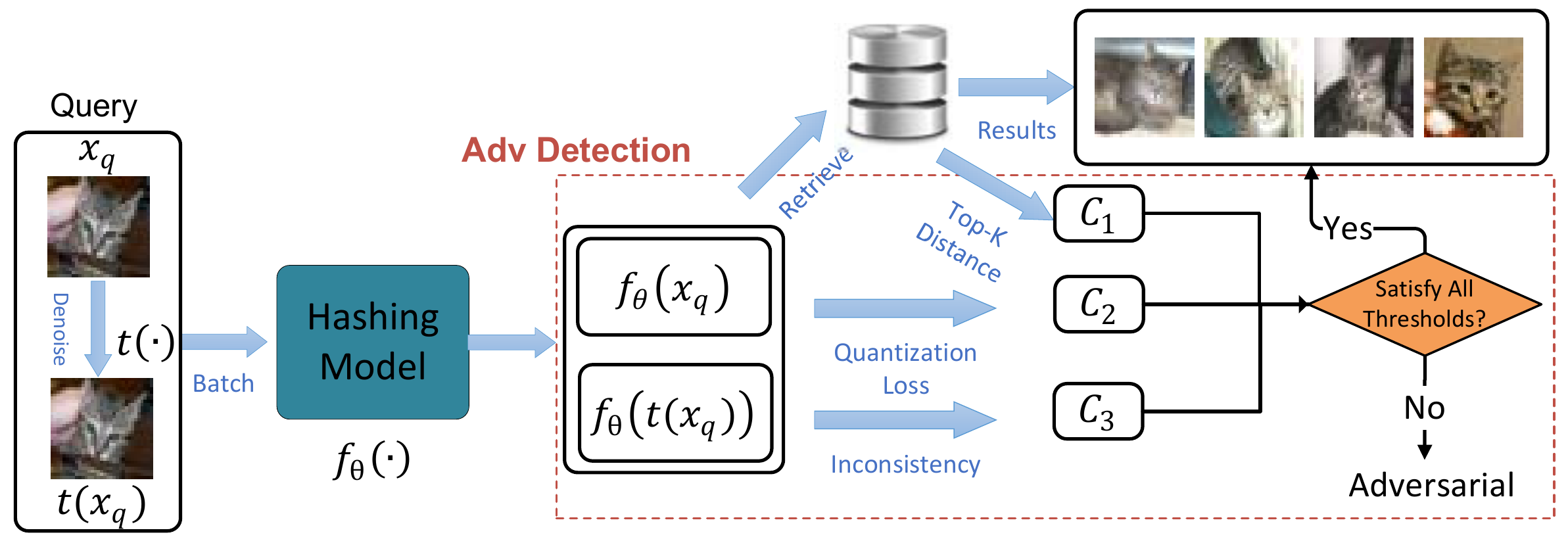}
\vspace{-0.05in}
\caption{The proposed detection framework: highlighted by the dash lines.}
\label{fig:framework}
\vspace*{-0.25in}
\end{figure*}

Starting from the untargeted attacks~\cite{tao}, we first theoretically deduce the hamming distance distribution from the adversarial image to other categories, which asymptotically approaches a Gaussian distribution. For targeted attacks, we discover an interesting adversarial behavior on the quantization loss: when the adversarial objective is to produce the same hash code of a targeted category~\cite{bai2020targeted,wang2021targeted}, it unintentionally brings the quantization loss close to zero. Thus, we first develop two thresholding methods that take hamming distance and quantization loss as the proxies. Then we combine the two criteria with a denoising-based detection to measure the disagreement between an input and its denoised transformation. We demonstrate that this combination can successfully defend against both \emph{gray-box} attackers, who have no prior knowledge of the detection method, and the strongest white-box attackers, who know the existence of the detection and can implement countermeasures. The overall framework is shown in Fig. \ref{fig:framework}.

The main contributions of this paper are summarized below. To the best of our knowledge, this is the first, unsupervised effort to defend against adversarial attacks in deep hashing. Based on the novel discoveries and analysis, we propose three criteria to unveil adversarial behaviors of targeted and untargeted attacks in the hamming space, and demonstrate their complementing relations against the strongest white-box attackers.
The extensive experiments on CIFAR-10, ImageNet, MS-COCO and NUSWIDE datasets show that the proposed method surpasses the state-of-the-art defenses by up to 23\% in detection rates with negligible computational overhead for real-time image queries.

\vspace{-0.05in}
\section{Preliminary}
This section illustrates the fundamentals of deep hashing and adversarial attacks.

\vspace{-0.05in}
\subsection{Deep Hashing}
Given a dataset of $N$ samples $X=\{x_1, x_2, \ldots, x_N\}$, $x_i \in \mathbb{R}^D$ and their corresponding labels $Y=\{y_1, y_2, \ldots, y_N\}$, $y_i \in \mathbb{R}^C$, where $x_i$ is the $i$-th sample and $y_{c,i}=1$ if the $i$-th image is associated with class $c$. Deep hashing learns a function $f_{\theta}(x)$ that maps the input image $x$ into a $K$-bit binary code $h(x)$ via a sign operation,
\begin{equation}
\small
\vspace{-0.02in}
h(x)=sign(f_{\theta}(x)) \in \{-1, +1\}^K,  \label{eq:hashingfunction}
\vspace{-0.02in}
\end{equation}
where $\theta$ are the parameters learned from minimizing the weighted combination of the similarity loss $\mathcal{L}_S$ and quantization loss $\mathcal{L}_Q$~\cite{dsh,hashnet,dch,zhu2016deep,lin2015deep,csq},
\begin{equation}
\small
\vspace{-0.05in}
\theta = \argmin\limits_{\theta} \mathcal{L}_S + \lambda \mathcal{L}_Q.  \label{eq:loss}
\vspace{-0.05in}
\end{equation}
$\mathcal{L}_S$ represents the hamming distance $D_h(h(x_i),h(x_j))$ between two images $x_i$ and $x_j$ with their similarity $s(y_i, y_j)$,
\begin{equation}
s(y_i, y_j) =
\begin{cases}
+1, & \mbox{if $y_i y_j^T > 0$ } \\
-1, & \mbox{otherwise}. \label{eq:similarity}
\end{cases}
\end{equation}
$\mathcal{L}_Q$ is the quantization loss to minimize the difference between the continuous output of $f_\theta(x)$ and its binary code $h(x)$. The objective is to minimize the hamming distance $D_h(h(x_i),h(x_j))$ between two samples $x_i$ and $x_j$ when they are similar, maximize the hamming distance when they are dissimilar, and meanwhile, represent the continuous $f_\theta(x)$ as binary codes. Both $D_h({h}_1, {h}_2)$ and $h(x)$ are non-differentiable regarding their inputs. A common technique is to use the differentiable form of $D_h({h}_1, {h}_2)$ noted as $\frac{1}{2}(K-{{h}_{1}}^T{h}_2)$ during backpropagation, where ${h}_1,{h}_2$ are the continuous floating point representation in $[-1, +1]$, and the binary hash codes $h(x)$ are represented by the continuous output of $f_\theta(x)$. The gap between such continuous and binary representations is considered as the quantization loss $\mathcal{L}_Q$, which is minimized in Eq. \eqref{eq:loss}.

Deep hashing consists of two main components, a \emph{database} and a \emph{model}. The database stores the images and their pre-computed hash codes. Given a query image $x$ with hash code $h(x)$, the system returns the top-$k$ images from the database which are $h(x)$'s $k$-nearest neighbors determined by hamming distance. The retrieval performance is calculated by the mean average precision (mAP), which is the ratio of images similar to $x$. In this paper, we base the hashing framework on the state-of-the-art method called Central Similarity Quantization (CSQ)~\cite{csq}. CSQ pre-determines the optimal hash codes based on the Hadamard matrix and randomly selects a set of hash codes with sufficient distances from each other as the hash centers from the Hadamard matrix (or from a random binary matrix if the Hadamard matrix is not available). Since different hashing techniques share the general objective of Eq. \eqref{eq:loss}, our defense applies to other techniques as well~\cite{dsh,hashnet,dch,zhu2016deep,lin2015deep}.

\vspace{-0.05in}
\subsection{Adversarial Attacks}

\textbf{Untargeted Attack}~\cite{tao} finds an adversarial image $x'$ by maximizing the hamming distance between the hash codes of adversarial examples and original images, subject to the $\mathcal{L}_\infty$ bound of $\epsilon$.
\begin{equation}\label{eq:untargeted}
\begin{aligned}
    \max \limits_{x'} D_h\bigl( h(x'),h(x)\bigr) \\
    \textrm{s.t.} \; \norm{x-x'}_{\infty} \leq \epsilon
\end{aligned}
\end{equation}
It works effectively to reduce the mAP by pushing the original image towards the furthest hamming distance in the hash space.

\textbf{Targeted Attack}~\cite{bai2020targeted,wang2021targeted,xiao2021you} attempts to minimize the hamming distance from $x'$ to the targeted hash code $h_t$ of a specific category,
\begin{equation}\label{eq:targeted}
\begin{aligned}
    \min \limits_{x'} D_h(h(x'), h_t) \\
    \textrm{s.t.} \; \norm{x-x'}_{\infty} \leq \epsilon
\end{aligned}
\end{equation}
Once the attacker has embedded the adversarial images in the database, targeted attacks enable image retrieval from a specific category upon user queries. For example, as illustrated in~\cite{xiao2021you}, the database could mistakenly return the advertisements of branded beer from the database upon the query of facial lotions.

\textbf{Attack Model.} Attackers can carry out both untargeted and targeted attacks. In particular, we consider two types of \emph{gray-box} and  \emph{white-box} attackers.  \emph{Gray-box} attackers have access to all the information including network architecture, weights and data, but are not aware of the existence of adversarial detection. The stronger \emph{white-box} attackers are aware of both the model function/parameters and the existence of the detection. So they implement different bypassing strategies as discussed in Section \ref{sec:exp}.

\vspace{-0.05in}
\section{Adversarial Behaviors in the Hamming Space}
Among a variety of artifacts left by adversarial images in classification networks, one of the most evident ``adversarial behaviors'' is from the softmax function~\cite{li2017adversarial,hendrycks2016baseline}. Due to the fast-growing exponentiation, it magnifies small changes in the logits~\cite{hendrycks2016baseline} and becomes overconfident in the presence of adversarial images by regularizing other categories~\cite{li2017adversarial}. In contrast to softmax, which makes the decision from a closed set of categories, hashing maps similar images into compact hamming balls in an open hamming space of $\{-1,1\}^K$. In this section, we define three criteria to identify adversarial behaviors in the hamming space.

\begin{figure*}[!ht]
\vspace*{-0.09in}
\centering
\hspace*{-0.05in}
\begin{subfigure}[b]{0.45\textwidth}
                \includegraphics[width=0.9\textwidth]{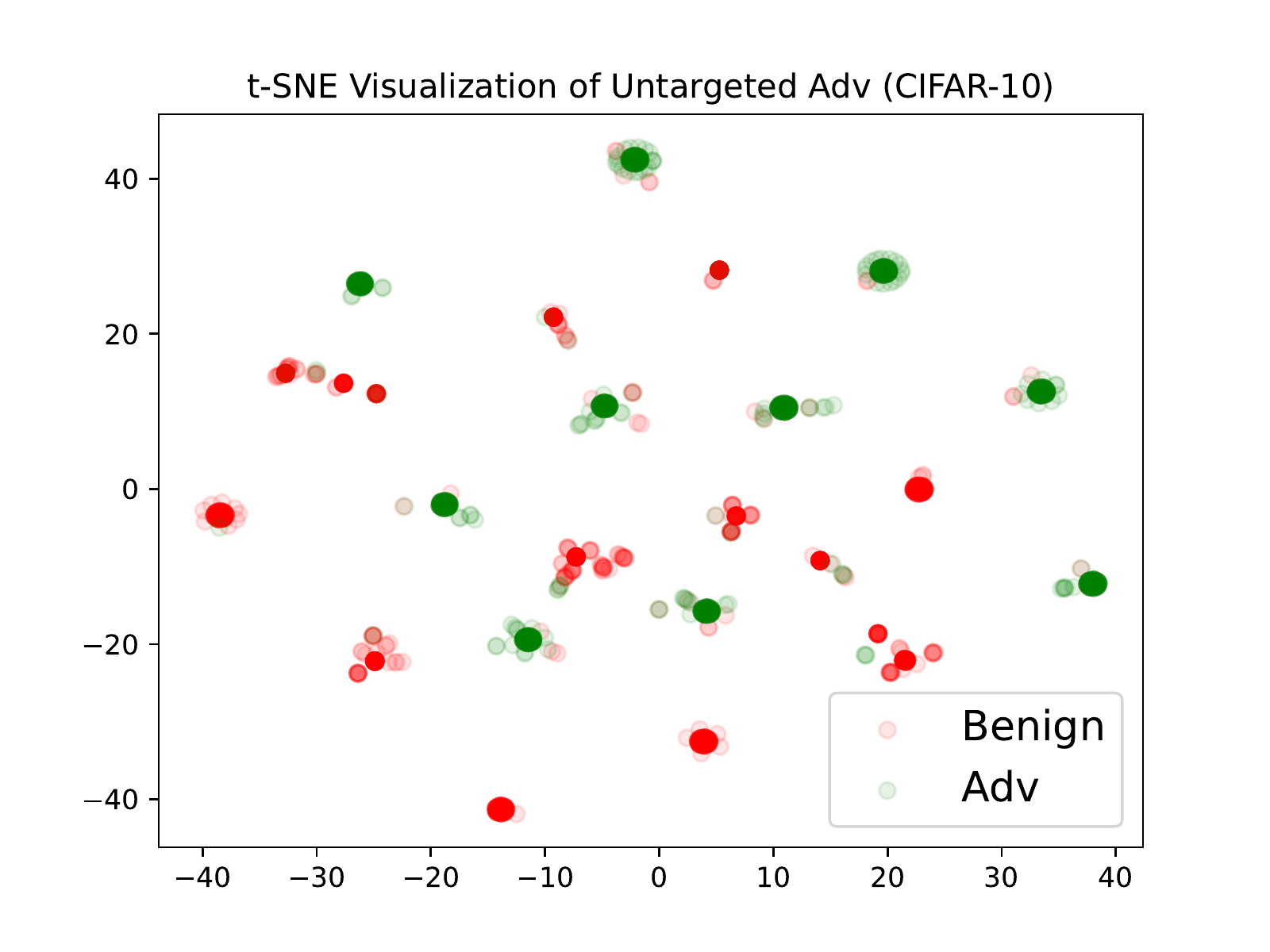}
                \vspace{-0.2in}
                \caption{}
\end{subfigure}
\hspace*{0.2in}
\begin{subfigure}[b]{0.45\textwidth}
                \includegraphics[width=0.9\textwidth]{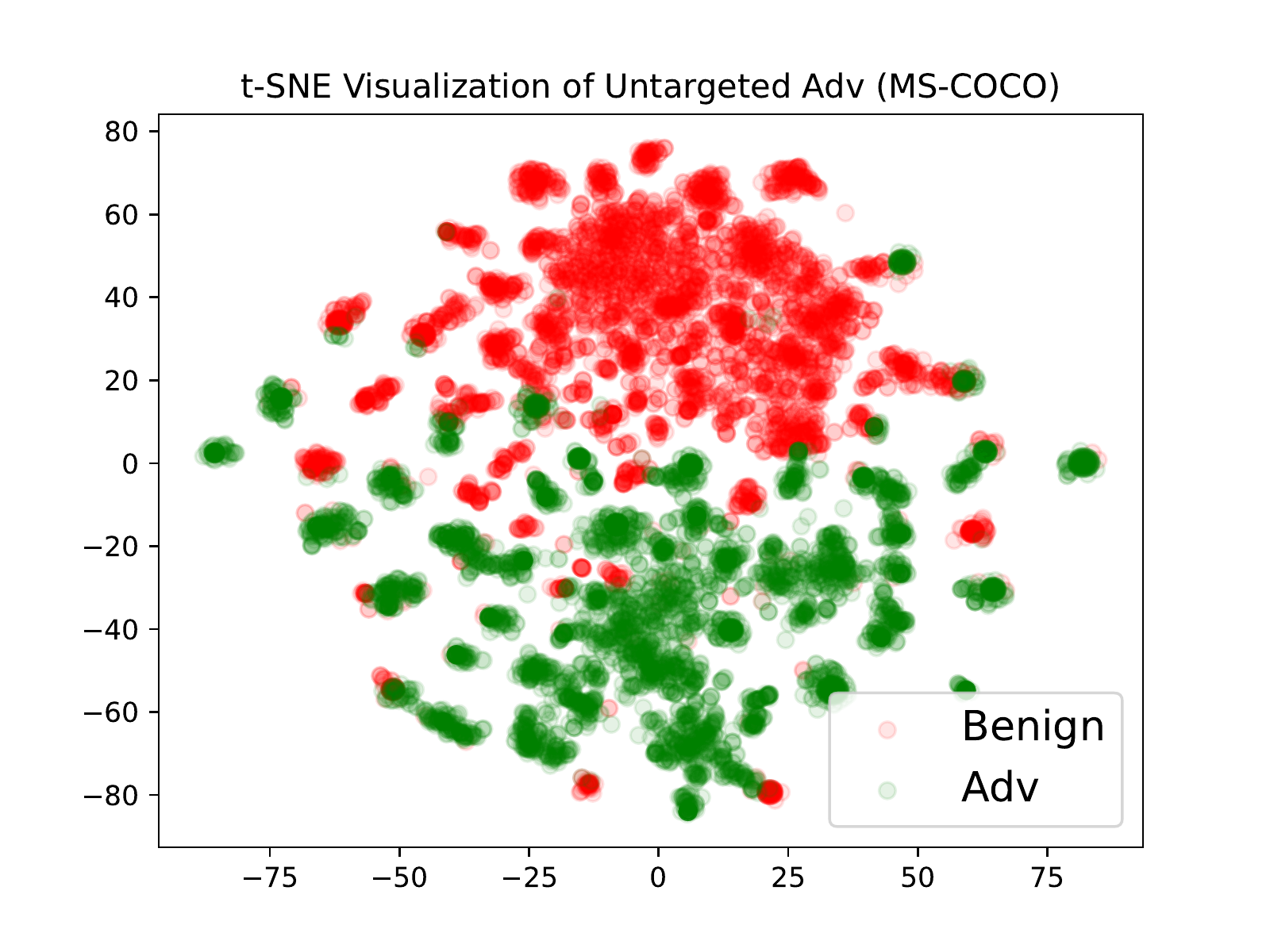}
                \vspace{-0.2in}
                \caption{}
\end{subfigure}%
%\hspace*{0.05in}
    \caption{t-SNE visualization of untargeted adversarial images vs. original images of different datasets (a) CIFAR-10. (b) MS-COCO.}
\label{fig:tsne_hashcode}
\vspace*{-0.2in}
\end{figure*}

\vspace{-0.05in}
\subsection{Detecting Untargeted Attacks ($C_1$)}

We start with untargeted attacks that maximize the hamming distance between $x'$ and $x$~\cite{tao}. Though such behavior is straightforward to discern, we seek a theoretical answer to the distribution of $h(x')$ when the attacking capacity is maximized.

\textbf{Assumption 1.} The network is capable of learning \emph{perfect} hash codes with the minimum intra-class distance (i.e., equals to zero) and maximum margin between each other.

If Assumption 1 holds, what is the hamming distance from the adversarial $h(x')$ to the rest of the hash codes? To answer this question, we first establish the distribution of the maximum inter-class distance as illustrated in the following Lemma.

\textbf{Lemma 1.} Given a number of $C$ classes in the $K$-bit hamming space with (ideally) compact hash codes, the inter-class hamming distance follows a Binomial Distribution of $\mathcal{X} \sim B(K, p)$, where $p=\frac{C}{2(C-1)}$ and $p \approx \frac{1}{2}$ when $C$ is large.
\begin{proof}
For all $C$ classes, consider only one bit location at a time. The maximum hamming distance is achieved when there is an equal number of $\frac{C}{2}$, $\{+1\}$ and $\{-1\}$ codes among the $C$ classes. The hamming distance between two bits is either $0$ or $1$. Thus, among $\binom{C}{2}$ selection of pairs, the probability that the hamming distance equals to $1$ is $(\frac{C}{2} \cdot \frac{C}{2})/\binom{C}{2} = \frac{C}{2(C-1)}$. Since all $K$ bits can be selected independently, the probability of the inter-class hamming distance between $h_i$ and $h_j$ equals to $d$ is,
\begin{equation}
\small
\vspace{-0.04in}
Pr \bigl( D_h(h_i, h_j ) = d \bigr) = \binom{K}{d} p^d (1-p)^{K-d}, \hspace{0.15in} p = \frac{C}{2(C-1)},   \label{prob_dist}
% \vspace{-0.00in}
\end{equation}
in which mean value is $K p$ and variance is $K p (1-p)$.
\end{proof}
From \emph{Lemma 1}, we can further deduce the next theorem.

\textbf{Theorem 1.} For untargeted attacks, the hamming distance from the adversarial image to any other classes follows a Gaussian distribution $\mathcal{N}\sim(K(1-p), Kp(1-p))$.
\begin{proof}
In the ideal situation, the untargeted attack maximizes the hamming distance from $D_h(h(x'), h(x_i))$ to $K$. Thus, for any other hash codes $h(x_j)$, the hamming distance is $D_h(h(x'), h(x_j)) = K - D_h(h_i, h_j)$, which is also a Binomial distribution with the mean of $K(1-p)$ and the same variance. When the hash bits $K$ is a large value, it can be approximated by a Gaussian distribution $\mathcal{N}\sim(K(1-p), Kp(1-p))$~\cite{ross2002first}.
\end{proof}

\textbf{Example.} When $K=64$ bits, and $C$ is large ($p \rightarrow \frac{1}{2}$), using the three-sigma rule, the confidence interval is $(K(1-p)-3 \sqrt{Kp(1-p)}, K(1-p)+3 \sqrt{Kp(1-p)})$. In other words, there is 99.73\% confidence that the hamming distance from an untargeted adversarial image to any other classes would be within the $[20,44]$ hamming distance interval with the mean of $K/2=32$, which is sufficiently distinguishable in the hamming space. Note that the above analysis serves a theoretical upper bound because achieving Assumption 1 is still an ongoing effort~\cite{csq,hoe2021one}. To see some examples, we visualize the t-SNE of untargeted adversarial images vs. benign images on CIFAR-10 and MS-COCO in Fig. \ref{fig:tsne_hashcode}. It is observed that despite of a few samples, the majority of the adversarial images are sufficiently distinguishable based on hamming distance. Hence, we formalize the first detection criterion.

\begin{figure*}[!ht]
\vspace*{-0.09in}
\centering
% \hspace*{-0.3in}
\begin{subfigure}[b]{0.33\textwidth}
                \includegraphics[width=1.0\textwidth]{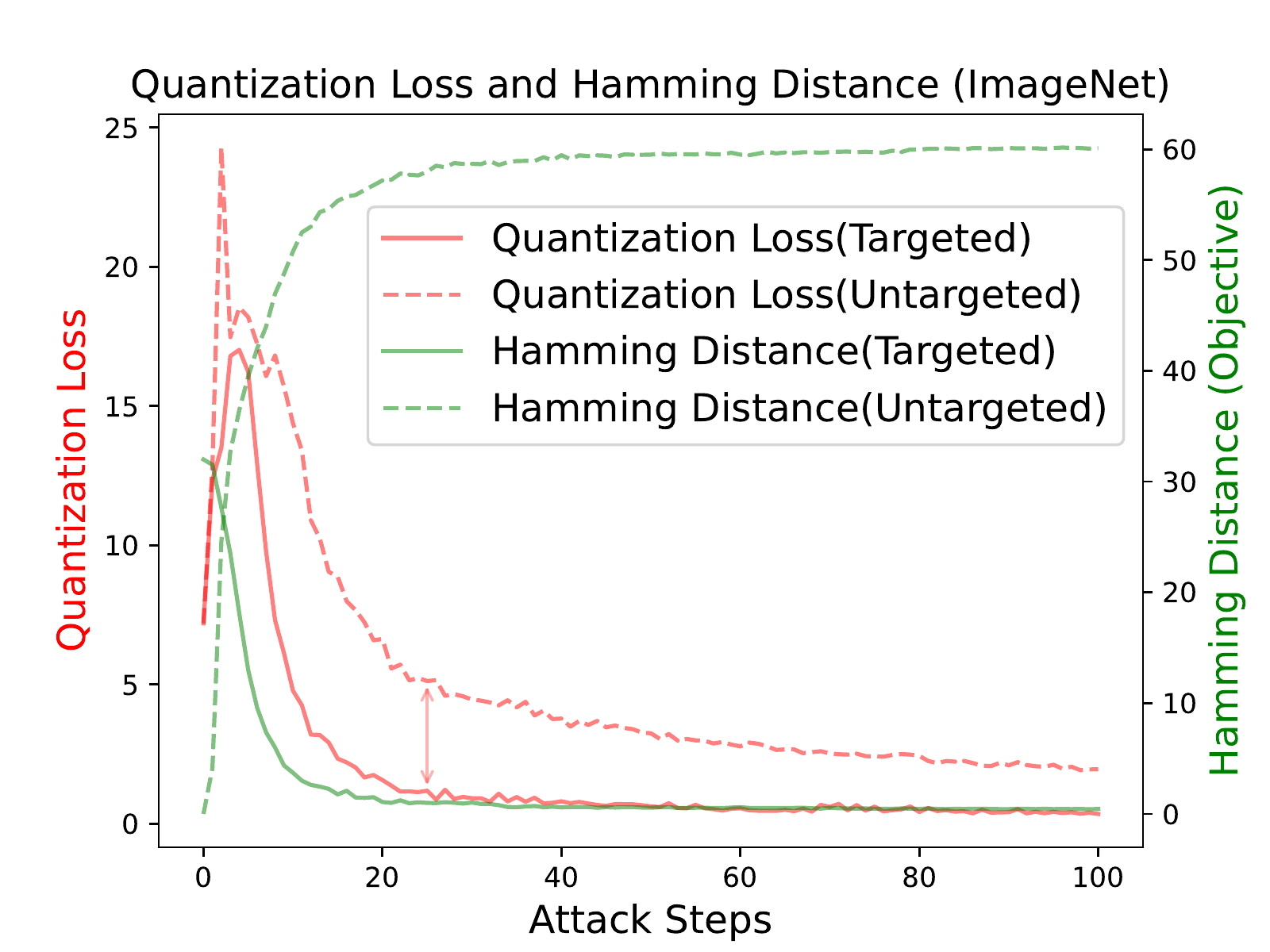}
                \vspace{-0.2in}
                \caption{}
\end{subfigure}
\begin{subfigure}[b]{0.33\textwidth}
                \includegraphics[width=1.0\textwidth]{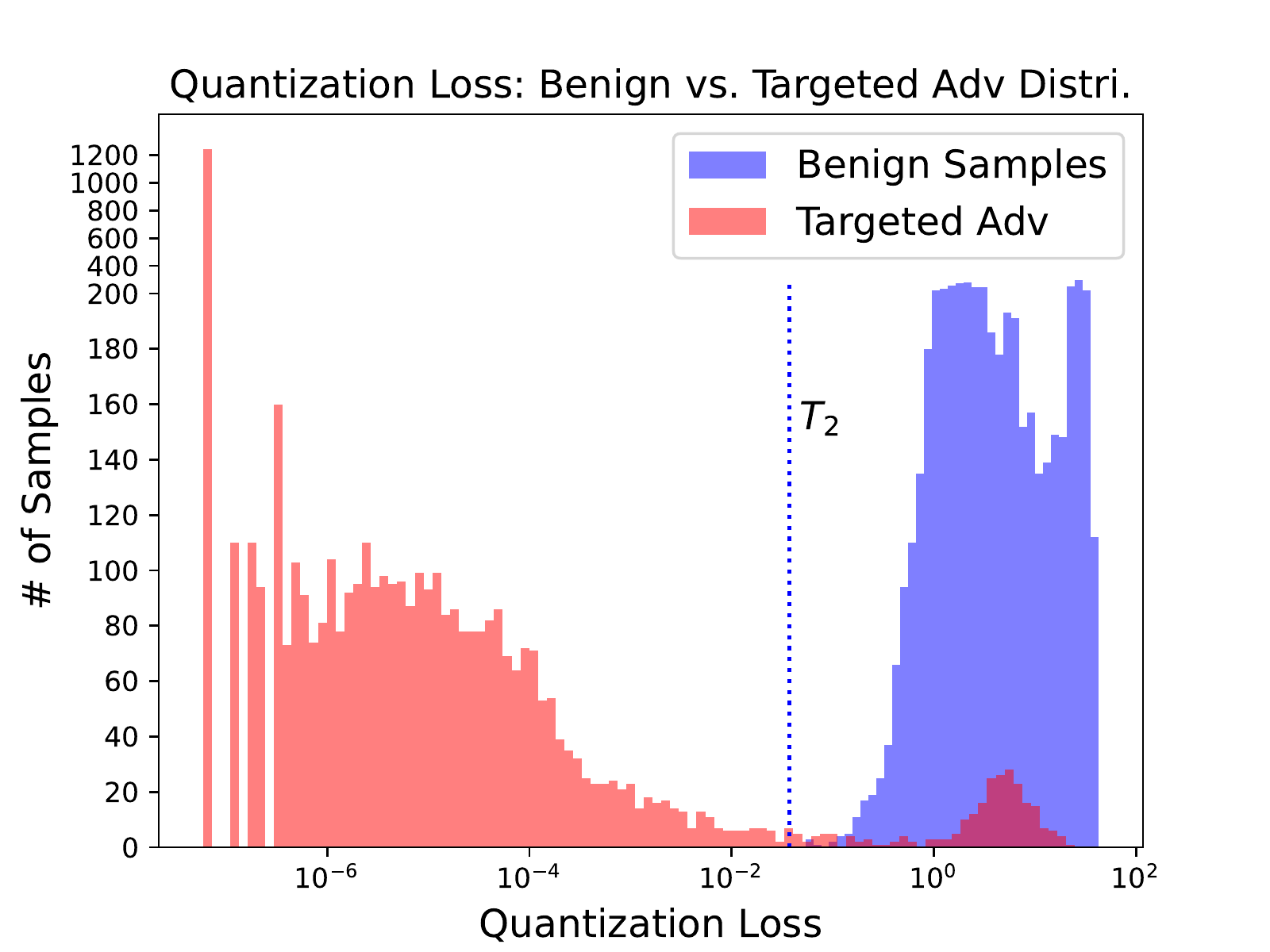}
                \vspace{-0.2in}
                \caption{}
\end{subfigure}%
\begin{subfigure}[b]{0.33\textwidth}
                \includegraphics[width=1.0\textwidth]{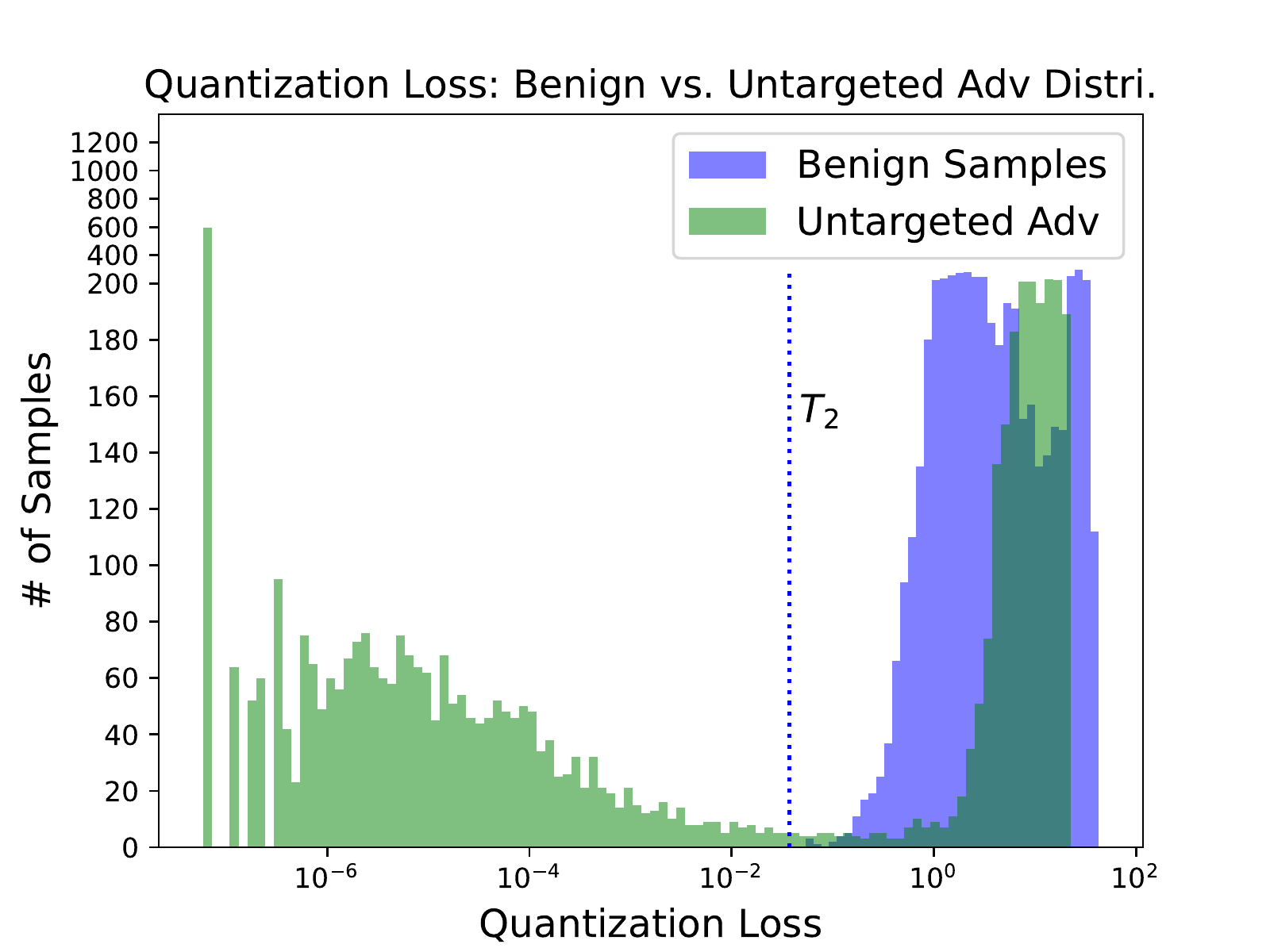}
                \vspace{-0.2in}
                \caption{}
\end{subfigure}%
%\hspace*{-0.2in}
\vspace*{-0.11in}
    \caption{Example of identifying targeted attacks based on quantization loss on ImageNet. (a) The quantization loss for targeted attacks concentrates around zero vs. the benign samples. (b) Targeted attacks push the quantization loss to zero compared to untargeted attacks. (c) 60\% of the untargeted attacks also concentrate around zero. }
\label{fig:criterion2_fig}
\vspace*{-0.15in}
\end{figure*}

\textbf{Criterion 1 (Hamming Distance).} For query $x$, collect the set of top-$k$ hash codes $\mathcal{H}_k$ and calculate the average hamming distance to $h(x)$.
\begin{equation}\label{eq:criterion_noman}
\small
\vspace{-0.05in}
{C}_1 = \frac{1}{|\mathcal{H}_k|} \sum_{h(x_k) \in \mathcal{H}_k} D_h \bigl(h(x_k), h(x)\bigr)
\vspace{-0.05in}
\end{equation}

${C}_1$ is the average hamming distance of the top-$k$ retrieval results, i.e., a scalar value and we can compare it with a threshold $\mathcal{T}_1$ calculated on benign samples. The computational process of ${C}_1$ follows the normal retrieval procedures using the top-$k$ hash codes. To detect targeted attacks, we develop the next criterion.

\vspace{-0.05in}
\subsection{Detecting Targeted Attacks (${C}_2$)}

While untargeted attacks attempt to induce a bit flip that makes $h(x')=-h(x)$, targeted attacks minimize the hamming distance between $h(x')$ and an arbitrary target code $h_t$ (e.g., such as computed from consensus voting~\cite{bai2020targeted} of a category). To find an appropriate metric to identify them, we have the following observation.

% \textbf{Observation 1.} For the quantization loss of benign images $\mathcal{L}_Q^b$ and targeted images $\mathcal{L}_Q^t$, the relation $\mathcal{L}_Q^b > \mathcal{L}_Q^t \approx 0$ holds.

\textbf{Observation 1.} For the quantization loss of benign images $\mathcal{L}_Q^b$ and targeted images $\mathcal{L}_Q^t$, the relation $\mathcal{L}_Q^b > \mathcal{L}_Q^t \approx 0$ holds.

To illustrate this observation, recall that the original targeted objective in Eq.~\eqref{eq:targeted} is not differentiable regarding the targeted binary code of $x'$. The implementation approximates via a continuous relaxation and the goal is to minimize the distance between the continuous output from the $\tanh(\cdot)$ function and the target code~\cite{hashnet}. As more gradient descent steps are taken, the quantization loss $\mathcal{L}_Q^t \rightarrow 0$, when their inter-distance is minimized. This is in close analogy with the adversarial images on softmax classifications while the targeted probabilities become overconfident~\cite{li2017adversarial}, and surprisingly, similar phenomenon is reflected on the quantization loss in deep hashing. In contrast, for all the benign samples, it is difficult to find the optimal model parameters to push $\mathcal{L}_Q^b$ towards zero during the training process. An example of ImageNet is shown in Fig.\ref{fig:criterion2_fig}(a) as the targeted attacks leave a distinguishable gap from benign samples. It is also interesting to compare with the quantization loss of untargeted attacks in Fig.\ref{fig:criterion2_fig}(a), which is larger than zero. This is because for untargeted attacks, finding an adversarial subspace that reduces the mAP to zero can be achieved before flipping all the bits, which is much easier than targeted attacks. Based on these observations, we develop the second detection criterion.

\textbf{Criterion 2 (Quantization Loss).} Calculate the $\bm{L}_p$ distance from the output of network $f_{\theta}(x)$ (logits before the sign function) and its hash code $h(x)$,
\begin{equation}
\small
C_2=\lVert h(x)-f_{\theta}(x)\rVert_p      \label{eq:criterion_qloss}
\end{equation}

$C_2$ is the quantization loss of query $x$. Here, we use the $\bm{L}_1$ distance ($p=1$) and obtain a threshold $\mathcal{T}_2$ on benign samples offline. Figs. \ref{fig:criterion2_fig}(b)(c) show the distribution of quantization loss between the adversarial and the benign images. As $C_2 \rightarrow 0$ for targeted attacks, we can see that using $\mathcal{T}_2$ can effectively identify most of the attacks; using $C_2$ also identifies about 60\% of the untargeted attacks,

%Note that the computation of $C_2$ also follows the normal retrieval procedures with $\mathcal{O}(K)$ computations

\vspace{-0.05in}
\subsection{Detecting Prediction Inconsistency ($C_3$)}

$C_1$ and $C_2$ alone are not sufficient. In principle, detection works by limiting the attacker's action space in a confined region. Perturbation can be generally treated as an artificial noise with high-frequency components~\cite{wang2020high}. Thus, a common approach is to apply local, non-local smoothing filters~\cite{xu2017feature}, auto-encoder denoiser~\cite{meng2017magnet}, color bit reduction~\cite{xu2017feature}, quantization~\cite{wangxiaofeng}, and measure the response sensitivity to the denoised images. The adversarial images are more prone to produce a different result, while the benign samples are less sensitive. These denoising operations reduce the entropy (randomness) and the input dimensions of adversarial space that the perturbations can act upon.

We extend this principle in deep hashing to formulate Criterion 3. Denote the transformation~\cite{xu2017feature,meng2017magnet,wangxiaofeng} as $t(\cdot)$. For query $x$, $C_3$ measures the hamming distance between a transformed $t(x)$ and $x$ based on the output before the sign function.

\noindent \textbf{Criterion 3 (Prediction Inconsistency).}
\begin{equation}\label{eq:criterion_denoiser}
\small
C_3 = D_h\bigl( f_{\theta}(t(x)),f_{\theta}(x) \bigr)
\end{equation}
In other words, $C_3$ quantifies the disagreement between the original and transformed inputs, which can be evaluated against a threshold $\mathcal{T}_3$ calculated offline on benign samples.
\vspace{-0.1in}
\subsection{Put Everything Together}
The overall detection combines the three criteria: given a query image $x$, we calculate $\{C_1,C_2, C_3\}$ and compare with thresholds $\{\mathcal{T}_1, \mathcal{T}_2, \mathcal{T}_3\}$. If (a) $C_1<\mathcal{T}_1$; (b) $C_2>\mathcal{T}_2$; (c) $C_3<\mathcal{T}_3$, the input is considered as benign; otherwise, if any of them is not satisfied, the input is rejected as an adversarial example. The computation time is bounded by $C_3$ since it requires two retrievals. To minimize the compute time, the system can combine the original query and its denoised copy into a batched query. In case the GPU has sufficient resources, it should have minimum overhead as discussed in Section~\ref{sec:compute_time}.

\vspace{-0.05in}
\section{Experiments} \label{sec:exp}

%For training, images are resized to 256 in the shorter axis, with the same ratio along the longer axis, then a random crop to $224 \times 224$ and a random horizontal flip are followed. All the images are normalized with $[0.485, 0.456, 0.406]$ mean value and $[0.229, 0.224, 0.225]$ std value, for all three color channels. For hash code extracting of database and query set, images are resized to $256$ in the same style training set does with, then a center crop to $224 \times 224$ is followed. %In figure ???, we show the distribution of $C_1$ on cifar-10 dataset, in both benign samples and untargeted samples. The distribution shows that the $C_1$ value in benign samples is much smaller than $C_1$ in untargeted samples. Since that, $C_1$ is used to detect untargeted adversarial attacks in deep hashing.

\vspace{-0.05in}
\subsection{Implementation}

We evaluate our mechanism on the CIFAR-10, ImageNet, MS-COCO and NUSWIDE datasets that are commonly used for deep hashing~\cite{dsh,hashnet,dch,zhu2016deep,lin2015deep,csq} and adopt CSQ~\cite{csq} with ResNet50~\cite{he2016deep} as the base model. The RMSProp optimizer~\cite{tieleman2012lecture} with learning rate $10^{-5}$ is used for training of 150 epochs. The weight of the quantization loss is set to $10^{-4}$. For four datasets, our trained models achieve mAP of 0.854, 0.883, 0.884, and 0.843, respectively.

We compare with several benchmarks originally designed for softmax classification:  Local Intrinsic Dimensionality (LID)~\cite{ma2018characterizing},  Median Smoothing (FS-Median), Non-local Means (FS-NLM)~\cite{xu2017feature}, FS-Adaptive~\cite{wangxiaofeng}, and MeanBlur~\cite{li2017adversarial}. FS-Adaptive uses the entropy of input as a metric to adaptively reduce the input space using scalar quantization and smoothing spatial filter. We select MeanBlur as the denoising technique for our method. True Positive Rate (TPR), False Negative Rate (FNR, Miss Rate), and Area-Under-Curve (AUC) are used as the evaluation metric, where adversarial examples are considered as Positive and benign samples are considered as Negative. Thus, a detected adversarial example is counted as a true positive, while a misidentified benign sample is counted as false positive.

% All detection methods are evaluated against both untargeted~\cite{tao} and targeted~\cite{bai2020targeted} PGD~\cite{pgd} and untargeted CW~\cite{carlini2017towards} attacks\footnote{The CW attack is initially designed for softmax classification and only a few logits are optimized. But for deep hashing, all hashing bits need to be optimized which rarely succeed for targeted CW attacks. Thus, we focus on the untargeted CW attack in deep hashing.}. For PGD, the step size is set to 1.0 and the $\bm{L}_{\infty}$ norm of perturbation $\epsilon$ is set to $8$, $16$, and $32$ with $100$ steps. For CW attack, the learning rate is set to $0.1$ and $0.01$ denoted as CW-a and CW-b with $500$ steps.

All detection methods are evaluated against both untargeted~\cite{tao} and targeted~\cite{bai2020targeted} deep hashing adversarial attacks (based on the PGD attack) and an untargeted deep hashing CW~\cite{carlini2017towards} attack\footnote{The CW attack is initially designed for softmax classification thus only a few logits are optimized. But for deep hashing, all hashing bits need to be optimized which rarely succeed for targeted CW attacks. Thus, we focus on the untargeted CW attack in deep hashing.}. The step sizes of the former two are set to 1.0 with $100$ steps, limited by an $\bm{L}_{\infty}$ norm of perturbation $\epsilon=8$. For CW attack, the learning rate is set to $0.01$ with $500$ steps. Additional experimental details and results are available in appendix.

\vspace{-0.05in}
\subsection{Detection of Gray-Box Attacks}

\begin{table}[!ht]
%\vspace{-0.1in}
\centering
\small
\begin{tabular}{@{}l l c c c @{}}
% \toprule&\multirow{2}[3]{*}{} & \multicolumn{2}{c}{PGD($\epsilon=8$)} & \multicolumn{1}{c}{CW-b}  \\
% \cmidrule(lr){3-4} \cmidrule(lr){5-6} \cmidrule(lr){7-8} \cmidrule(lr){9-9} \cmidrule(lr){10-10}
% & &Untgt & Tgt &Untgt & Tgt & Untgt & Tgt  & Untgt & Untgt\\
&&Untgt~\cite{tao}& Tgt~\cite{bai2020targeted} & Untgt CW~\cite{carlini2017towards}  \\

\midrule
\multirow{6}{1mm}{\begin{sideways}\parbox{17mm}{CIFAR-10}\end{sideways}}

& LID~\cite{ma2018characterizing} &  14.90 & \textcolor{blue}{\textbf{4.00}}& 12.16\\
%& KDE& xxx& xx& x.x& x.x & x.x & x.x\\
% & FS-BitDepth~\cite{xu2017feature}& 0.0000& 0.0000& 0.0570& 0.0000& 0.0000& 0.0000& 0.0250& 0.4790\\
%\cline{2-10}
& FS-Median~\cite{xu2017feature}&  13.40& 54.30 & \textcolor{blue}{\textbf{2.40}}\\
& FS-NLM~\cite{xu2017feature}& 18.80& 18.10 & 14.41\\
& FS-Adaptive~\cite{wangxiaofeng}& 54.00& 48.30 & 52.55\\
& MeanBlur~\cite{li2017adversarial}&  \textcolor{blue}{\textbf{11.60}}& 26.90& \textcolor{blue}{\textbf{2.40}}\\

& \textbf{UMCD} (Ours)&  \textcolor{red}{\textbf{7.50}} & \textcolor{red}{\textbf{0.40}} & \textcolor{red}{\textbf{0.15}}\\

\hline

\multirow{6}{1mm}{\begin{sideways}\parbox{17mm}{ImageNet}\end{sideways}}
& LID~\cite{ma2018characterizing} &21.78 & 6.40 & 75.95\\
%& KDE& xxx& xx& x.x& x.x & x.x & x.x\\
% & FS-BitDepth~\cite{xu2017feature} & 0.0104& 0.0018& 0.0004& 0.0000& 0.0000& 0.0000& 0.5649& 0.9858\\
%\cline{2-10}
& FS-Median~\cite{xu2017feature}& 8.54& 19.20& 1.41\\
& FS-NLM~\cite{xu2017feature}&  2.62& \textcolor{blue}{\textbf{3.40}}& 8.02\\
& FS-Adaptive~\cite{wangxiaofeng}&4.05& 4.30& 13.68\\
& MeanBlur~\cite{li2017adversarial}& \textcolor{blue}{\textbf{0.76}}& 4.54 & \textcolor{blue}{\textbf{0.94}}\\

& \textbf{UMCD} (Ours)& \textcolor{red}{\textbf{0.42}}& \textcolor{red}{\textbf{0.34}} &\textcolor{red}{\textbf{0.47}}\\

\hline

\multirow{6}{1mm}{\begin{sideways}\parbox{17mm}{MS-COCO}\end{sideways}}
& LID~\cite{ma2018characterizing}&3.24 & \textcolor{blue}{\textbf{1.20}}&54.59\\
& FS-Median~\cite{xu2017feature}& 0.26& 7.10& \textcolor{red}{\textbf{0.00}}\\
& FS-NLM~\cite{xu2017feature}&  \textcolor{red}{\textbf{0.00}}& 1.72 & 0.08\\
& FS-Adaptive~\cite{wangxiaofeng}& \textcolor{red}{\textbf{0.00}}& 2.84 & 0.08\\
& MeanBlur~\cite{li2017adversarial}& \textcolor{red}{\textbf{0.00}}& 2.12& \textcolor{red}{\textbf{0.00}}\\
& \textbf{UMCD} (Ours)& \textcolor{red}{\textbf{0.00}}& \textcolor{red}{\textbf{1.12}} & \textcolor{red}{\textbf{0.00}}\\

\hline

\multirow{6}{1mm}{\begin{sideways}\parbox{17mm}{NUSWIDE}\end{sideways}}
& LID~\cite{ma2018characterizing}& 25.34 & \textcolor{red}{\textbf{0.08}}&47.96\\
& FS-Median~\cite{xu2017feature}& 3.52& 16.33 & \textcolor{red}{\textbf{0.00}}\\
& FS-NLM~\cite{xu2017feature}&\textcolor{red}{\textbf{0.00}}& 2.90& \textcolor{red}{\textbf{0.00}}\\
& FS-Adaptive~\cite{wangxiaofeng}&  \textcolor{red}{\textbf{0.00}}& 4.57 & \textcolor{red}{\textbf{0.00}}\\
& MeanBlur~\cite{li2017adversarial}& 0.19& 4.29 & \textcolor{red}{\textbf{0.00}}\\

& \textbf{UMCD} (Ours)&  \textcolor{red}{\textbf{0.00}}& \textcolor{blue}{\textbf{1.29}} & \textcolor{red}{\textbf{0.00}}\\
\bottomrule
\end{tabular}
%\vspace{-0.05in}
\caption{A comparison with state-of-the-arts methods of \emph{Detection Miss Rate} (False Negative Rate) against adversarial attacks for deep hashing, when allowing 5\% FPR on benign samples. Lower miss rate is better. Top two numbers of each column are \textbf{bolded}, with the best in \textcolor{red}{\textbf{red}} and the second in \textcolor{blue}{\textbf{blue}}. }%CW is implemented by ourselves since no CW attack is available on deep hashing yet.}
\label{tab:tpr}
\vspace{-0.1in}
\end{table}

\textbf{Our Method.} Table~\ref{tab:tpr} shows the detection miss rate of different methods when we fix the FPR at 5\%. The proposed method can make robust detection of targeted attacks with less than 1\% miss rate in the most cases. Compared to the SOTA benchmarks, our method is $2.13\%$ to $23.44\%$  better than other methods on average. Untargeted attacks are relatively easier to detect, so the overall miss rates are lower than targeted attacks. Our method can still improve the baselines by $1.15\%$-$14.33\%$.

\textbf{Compare with LID.} Although LID performs well on most of the targeted attacks, it does not generalize to the CW attack. Furthermore, different from the softmax networks that the last few layers often offer better detection~\cite{ma2018characterizing}, LID is quite sensitive to which layers should those features be extracted in deep hashing, which adds extra configuration overhead. In sum, LID is not always effective against all types of attacks.

% (Rewrite) On the other hand, it is interesting to see that the reduction of color bit depth (FS-BitDepth) does not provide defense against PGD attacks at all as most of the detection rates are around zero, but it performs contrastively better on CW-b. This is because bit depth reduction is only effective to filter out small noise/perturbations from the CW attacks, but not for large noise from the PGD attacks. Once the small perturbations from CW attacks are removed, the adversarial input and its denoised copy would yield large difference and get detected. This is also validated from CW-a and CW-b: since CW-b has lower learning rate than CW-a, it results in smaller perturbations and the detection rates are much higher than CW-a. In sum, LID and Color Bit Depth reduction are only effective against a single type of attacks.

\textbf{Compare with Spatial Denoising Methods.}  LID does not take advantage of the spatial information like the rest four benchmarks using non-local mean, median, etc. As shown in Table~\ref{tab:tpr}, though they generally have less than 10\% miss rates on untargeted attacks, there is a 0.5-17.8\% gap in detecting targeted attacks. Such gaps can be explained by the attack mechanisms as targeted  attacks take more gradient steps. This lands the image deep into the adversarial space, which is more robust to pixel-level modifications such as denoising~\cite{hu2019new,xiao2021you}. However, our method provides an extra layer of defense from $C_2$ to specifically monitor the value of quantization loss as targeted attacks bring it to zero, thereby complementing $C_3$ when targeted attacks push the inputs deep into the adversarial space.

% \textbf{Compare with Spatial Denoising Methods.}  Reduction of bit depth does not take advantage of the spatial information like the rest four benchmarks using non-local mean, median, etc. As shown in Table~\ref{tab:tpr}, though they generally have over 0.9 detection rates on untargeted attacks, there is a 10-20\% gap in detecting targeted attacks. Such a gap can be explained by the attack mechanisms as targeted PGD attacks take more gradient steps. This lands the image deep into the adversarial space, which is more robust to pixel-level modifications such as denoising~\cite{hu2019new,xiao2021you}. However, our method provides an extra layer of defense from $C_2$ to specifically monitor the value of quantization loss as targeted attacks bring it to zero, thereby complementing $C_3$ when targeted PGD attacks push the inputs deep into the adversarial space.

%Overall, our method works well with combination of deep hashing adversarial behaviors $C_1$, $C_2$ and a denoising-based criterion $C_3$. It perform especially well on targeted attacks, where might exist the complementarity between criterion $C_2$ designed for targeted attacks and the denoising-based criterion $C_3$. That leads to our question: Does this complementary property help the robustness when it comes to white-box scenario? We answer this question in the next subsection.

\begin{table*}[!ht]
\vspace{-0.08in}
\centering
\small
\hspace{-0.08in}\begin{tabular}{l  c c  c c  c c c c c c @{}}
\toprule

& \multicolumn{2}{c}{CIFAR-10} &  \multicolumn{2}{c}{ImageNet} &  \multicolumn{2}{c}{MS-COCO} &  \multicolumn{2}{c}{NUSWIDE} & \multicolumn{2}{c}{\emph{Avg. Gain}}  \\
\cmidrule(lr){2-3} \cmidrule(lr){4-5} \cmidrule(lr){6-7} \cmidrule(lr){8-9} \cmidrule(lr){10-11}
&Untgt &Tgt    &Untgt &Tgt    &Untgt &Tgt    &Untgt &Tgt     &Untgt &Tgt\\ %\hline

\midrule
$C_3$ Alone   & 0.8160 & 0.7460 & 0.9110& 0.8338 & 0.9954 &0.9076& 0.9757& 0.8904    & -- & --    \\
$C_1+C_3$  & 0.9870 & 0.7580 & 0.9522& 0.8460 & 0.9966 &0.9098& 0.9757& 0.8904   &\emph{0.0533}& \emph{0.0066}\\%&\emph{0.1061} & 0.0074  \\
$C_2+C_3$   & 0.8170 & 0.9830  & 0.9504& 0.9828 & 0.9992 &0.9784& 0.9961& 0.9847  &\emph{0.0161} & \emph{0.1377}\\ %& \emph{0.0450} & \emph{0.1688}  \\ \hline
$C_1+C_2+C_3$  & 0.9880 & 0.9950  & 0.9916& 0.9956 & 0.9992 &0.9784& 0.9961& 0.9847  &\emph{0.0692}& \emph{0.1439}\\ %& 0.1263 & 0.1762 \\
 \hline
\end{tabular}
\caption{Ablation study: detection rates of different combinations ($\epsilon=32$)}  \label{tab:ablation}
\vspace{-0.15in}
%\vspace{-0.15in}
\end{table*}

\vspace{-0.05in}
\subsection{Detection of White-box Attacks}
Next, we demonstrate the detection of white-box attackers, who know the existence of our detection and conduct countermeasures accordingly. The attacker adopts \emph{backward pass differentiable approximation}~\cite{athalye2018obfuscated} to estimate the gradients and develops different strategies against $C_1$, $C_2$ and $C_3$:

\textbf{Against $\bm{C_1}$}. $C_1$ relies on the hamming distance between hash codes to detect outliers. Thus, an effective evasion is to drive the adversarial examples into the neighborhoods of benign images, e.g., generating the same binary hash codes of certain targeted images $h_t$:
\begin{equation}
\small
\mathcal{L}_1 = \underbrace{D_h(f_\theta(x'), h_t)}_\textrm{adv loss}
\end{equation}

\textbf{Against $\bm{C_2}$}. $C_2$ detects near-zero quantization loss by accessing the logits before the sign function. To bypass this detection, the attacker aims to maximize the quantization loss, which amortizes the adversarial behavior identified from $C_2$.
\begin{equation}
\small
\mathcal{L}_2 = - \underbrace{\norm{h(x')-f_{\theta}(x')}_1}_\textrm{quantization loss}
\label{eq:adapt_qloss_only}
\end{equation}

\textbf{Against $\bm{C_3}$}. $C_3$ detects the disagreement between $f_\theta(x')$ and the denoised copy $f_\theta(t(x'))$. The attacker minimizes such difference by enforcing distance between $f_\theta(x')$ and $f_\theta(t(x'))$ to be small,
\begin{equation}
\small
\mathcal{L}_3 =  \underbrace{D_h(f_\theta(t(x')), f_\theta(x'))}_\textrm{denoised adv loss}
\label{eq:adapt_denoise}
\end{equation}
We use MeanBlur~\cite{li2017adversarial} as the transformation $t(\cdot)$ here. By combining them, the white-box attacker constructs a joint optimization objective,
\begin{equation}
\small
\min \limits_{x'} \mathcal{L} = \mathcal{L}_1 + \lambda_1 \mathcal{L}_2 + \lambda_2 \mathcal{L}_3
% \min\limits_{x'} \underbrace{D_h(f_\theta(x'), h_t)}_\textrm{adv loss} - \underbrace{\lambda_1||h(x')-f_{\theta}(x')||_1}_\textrm{quantization loss} + \underbrace{\lambda_2 D_h(f_\theta(t(x')), f_\theta(x'))}_\textrm{denoised adv loss}.
\label{eq:adapt_combine}
\end{equation}

\begin{figure*}[!t]
% \vspace*{-0.09in}
\centering
\hspace*{-0.15in}
\begin{subfigure}[b]{0.5\textwidth}
    \includegraphics[width=0.95\textwidth]{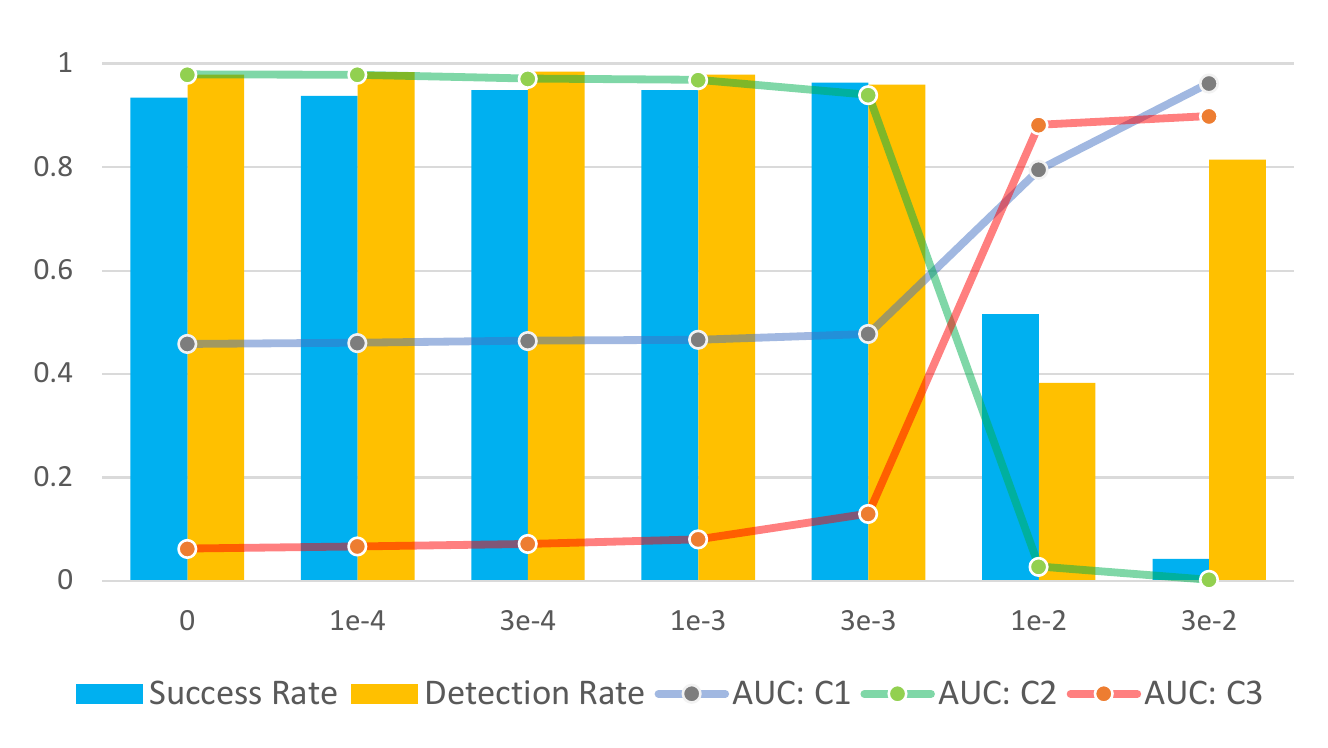}
    \caption{$\lambda_2=0.3$}
% \vspace{-0.2in}
\end{subfigure}
\begin{subfigure}[b]{0.5\textwidth}
    \includegraphics[width=0.95\textwidth]{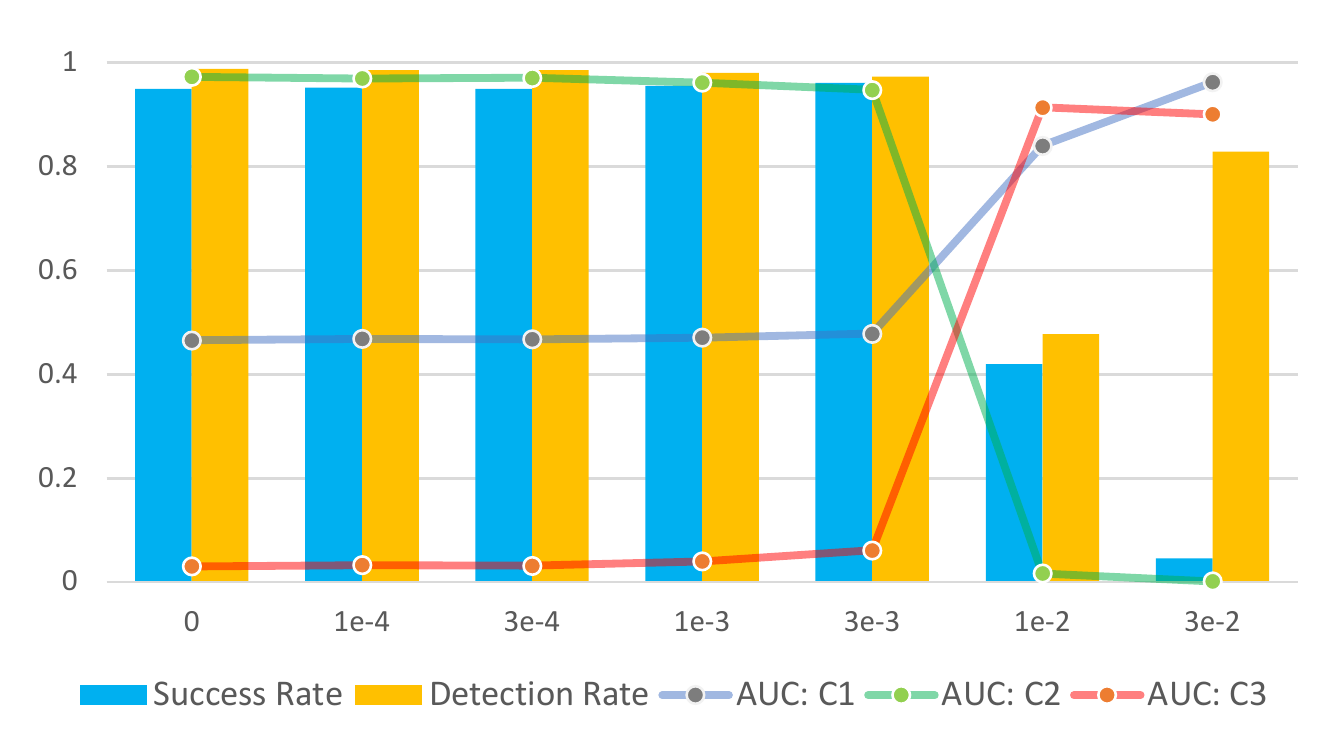}
    \caption{$\lambda_2=3$}
% \vspace{-0.2in}
\end{subfigure}
% \includegraphics[width=0.5\textwidth]{images/new_white_box/1.pdf}
% \vspace{-0.2in}

    \caption{White-box attack results. x-axis is the $\lambda_1$ value and y-axis is the percentage. Once the white-box attacks achieve a lower detection rate at $\lambda_1=$1e-2, the success rate of generating such adversarial examples also drops significantly. The three detection criteria compensate each other by observing the AUC values: when $C_1,C_3$ are low, $C_2$ is high and vice versa. }
\label{fig:whitebox}
\vspace*{-0.15in}
\end{figure*}

\textbf{Detection of White-box Attacks.} Optimizing \eqref{eq:adapt_combine} turns out to be quite difficult by finding $\lambda_1$, $\lambda_2$ that breach all three criteria. We demonstrate the best effort in Fig.~\ref{fig:whitebox} that fixes $\lambda_2$ to $0.3$ and $3$ and adjusts $\lambda_1$ from $0$ to $0.03$. The best case is when $\lambda_1$=$0.01$, the detection rate has been lowered to around 0.4 (the yellow bars). However, the number of adversarial examples that can successfully optimize \eqref{eq:adapt_combine} is also decreased to around 45\% (the blue bars of success rate). Hence, although the strongest white-box attackers still have some chances, our detection has successfully confined the adversarial space by enlarging the attacker's efforts. It is also interesting to see that different criteria form compensating relations as indicated by the AUC values. When $C_2$ (green curve) declines, $C_1, C_3$ quickly rise and vice versa. This relation is further validated by the ablation study next.

\vspace{-0.07in}
\subsection{Ablation Study}
We present the ablation study to quantify the contribution of each criterion. We use $C_3$ alone as the baseline and add $C_1$ and $C_2$ with their averaged gain shown in Table~\ref{tab:ablation}. The result is consistent with the defense objectives as the addition of $C_1$ and $C_2$ helps improve the detection rates of the untargeted and targeted attacks by $\bm{0.0533}$ and $\bm{0.1377}$, respectively. Meanwhile, $C_1$ and $C_2$ contribute almost independently on the overall detection, e.g., $C_1+C_3=\bm{0.0533}$ plus $C_2+C_3=\bm{0.0161}$ is equal to $C_1+C_2+C_3 = \bm{0.0692}$ for untargeted attacks and the same also holds for targeted attacks. This validates that all three criteria act as indispensable parts in the detection.

\vspace{-0.05in}
\subsection{Computational Time} \label{sec:compute_time}
\begin{figure*}[!t]
\vspace*{-0.09in}
\centering
\hspace*{-0.15in}
\begin{subfigure}[b]{0.45\textwidth}
                \includegraphics[width=1.0\textwidth]{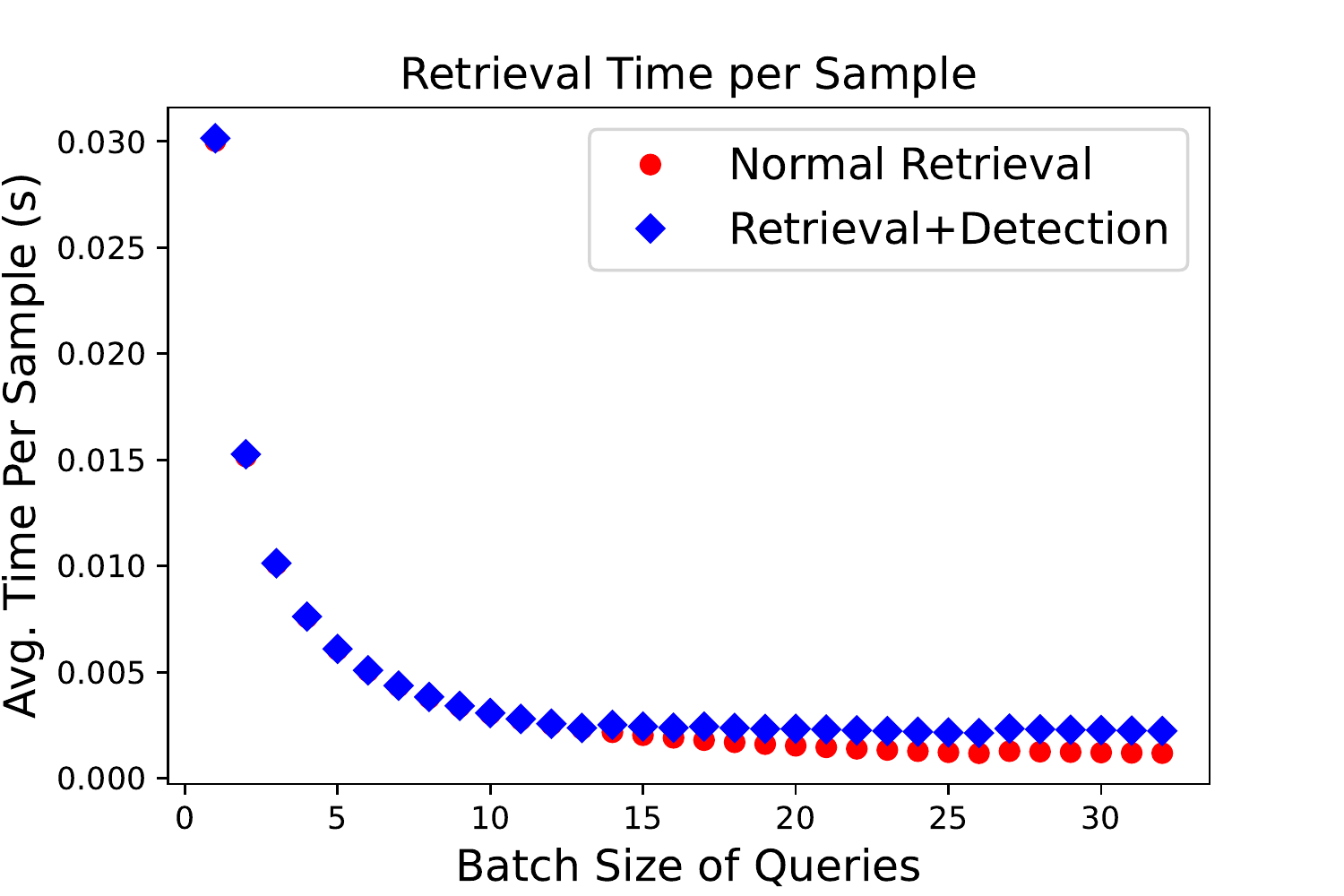}
                \vspace{-0.2in}
                \caption{}
\end{subfigure}
\begin{subfigure}[b]{0.45\textwidth}
                \includegraphics[width=1.0\textwidth]{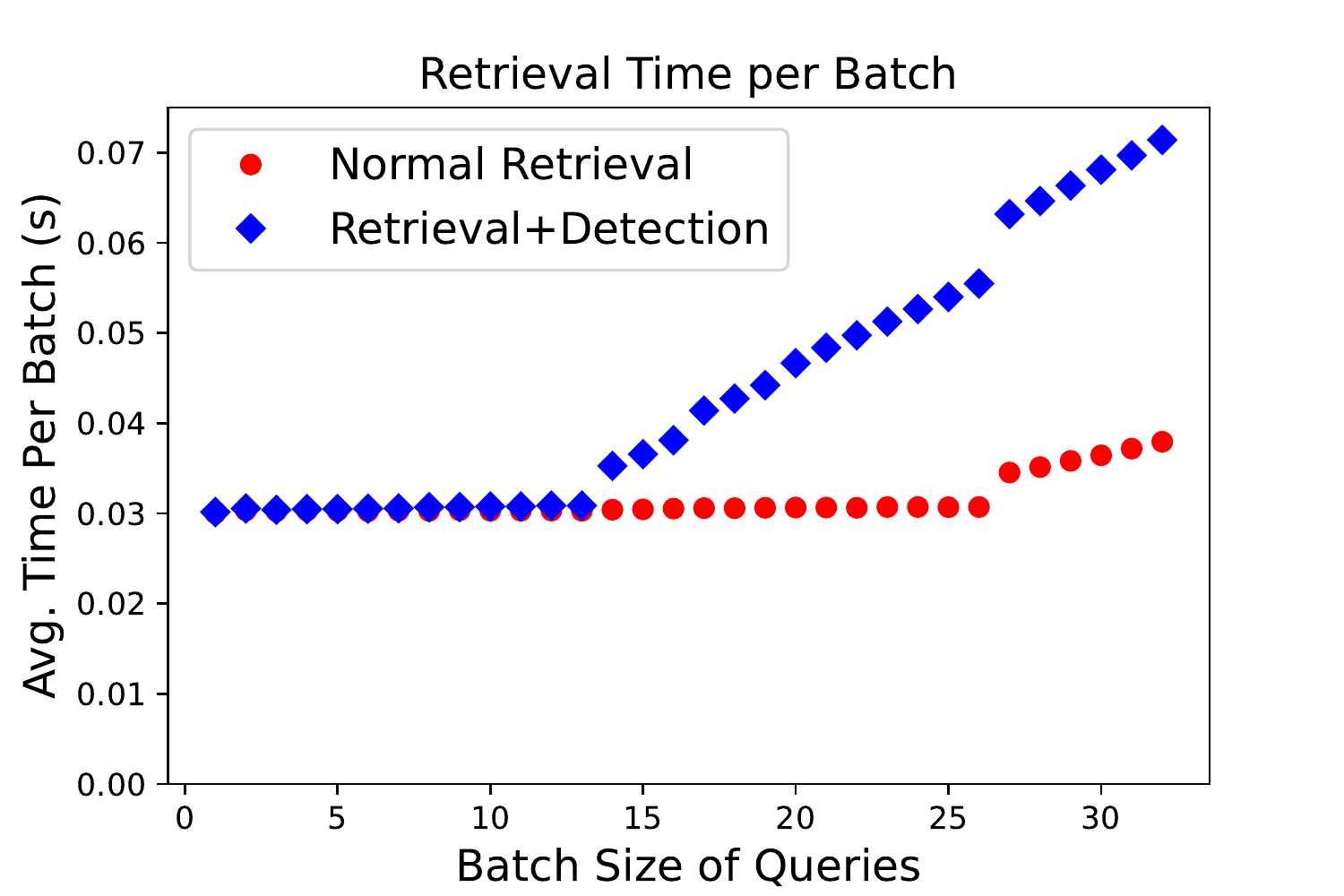}
                \vspace{-0.2in}
                \caption{}
\end{subfigure}%
%\hspace*{-0.2in}
\vspace*{-0.11in}
    \caption{Computation time of different batch size: a) per sample; b) per batch. }
\label{fig:time_cost}
\vspace*{-0.15in}
\end{figure*}

Finally, we evaluate the computational overhead of the detection mechanism. In practice, the system can accumulate queries into a batch to enhance the utilization of GPU resources and reduce cost. Fig.~\ref{fig:time_cost} shows the average retrieval time per sample/batch. First, it is observed that the average time per sample is under 50 ms and further reduced as we increase the batch size. Once the batch size is small, detection introduces negligible overhead because the GPU is underutilized; as the batch size increases, an additional retrieval from the denoised copy in $C_3$ enlarges the gap between normal retrieval since the GPU resources have been fully utilized. Thus, our detection introduces minimum overhead when the system accumulates relatively small batch and responds to queries in real-time.

\vspace{-0.05in}
\section{Related Work}

\vspace{-0.05in}
\subsection{Deep Hashing}

Image retrieval uses nearest neighbor search to return the semantically related images of query inputs. Traditionally, it relies on hand-crafted visual descriptors to reduce the computational cost of similarity measure~\cite{oliva2001modeling,dalal2005histograms}. Powered by deep learning, end-to-end hash learning improves the performance to a new level~\cite{dsh,hashnet,dch,zhu2016deep,lin2015deep,csq}. They use the similarities between image pairs to train deep hashing models in a supervised manner by transforming the high-dimensional images into compact hash codes, on which neighboring search can be efficiently performed based on hamming distance. To convert the continuous outputs into discrete binary codes, common approaches use continuous relaxation such as sigmoid or hyperbolic tangent functions to approximate the discrete binary thresholding~\cite{dsh,hashnet,dch,zhu2016deep,lin2015deep,csq}. Our work exploits the adversarial behaviors originated from this approximation process, thus can be applied to a variety of deep hashing models.

\vspace{-0.05in}
\subsection{Adversarial Attacks}

Deep neural networks are known to be vulnerable to the non-perceptible perturbations~\cite{szegedy2013intriguing}. The Fast Gradient Sign Method (FGSM)~\cite{fgsm} generates perturbations in the direction of the signed gradient to maximize the loss function in one-shot computation. The Basic Iterative Method (BIM)~\cite{bim} and Projected Gradient Descent (PGD)~\cite{pgd} take iterative steps (from random initialization) to achieve higher attack success. There are several other variants~\cite{carlini2017towards,athalye2018obfuscated,papernot2017practical,tramer2017space}, e.g., the CW attack aims at minimizing the perturbations to evade detection.

By using the deep learning backends, deep hashing inherits the vulnerability from neural networks. With some slight adaptation, recent works have shown that adversarial attacks can also mislead image retrieval systems~\cite{tao,xiao2020evade,bai2020targeted,xiao2021you,wang2021targeted}. The attacks can be generally categorized into \emph{untargeted} and \emph{targeted attacks}. \emph{Untargeted attacks} divert the query away from the correct results, which make the system retrieve irrelevant images or simply nothing. \cite{tao} proposes an untargeted attack to maximize the hamming distance between adversarial and benign samples. \cite{li2021qair,chen2021dair} craft adversarial examples based on iterative retrievals from a black-box model. \cite{xiao2020evade} hides private images in the database into a non-retrievable subspace by minimizing the number of samples around the private images. \emph{Targeted attacks} make the systems return images from a targeted category, different from the inputs. \cite{bai2020targeted,wang2021targeted} minimize the average hamming distance between the adversarial examples and a set of images with a target label. \cite{xiao2021you} enhances targeted transferability to a black-box model via injecting random noise into the adversarial generation. Our work defends against both untargeted and targeted attacks in deep hashing.

\vspace{-0.05in}
\subsection{Adversarial Defenses}
Most of the defense mechanisms are based on softmax classification. As proactive measures, gradient masking~\cite{papernot2016distillation} and adversarial training~\cite{pgd,croce2020reliable,shafahi2019adversarial,wong2020fast} learns a robust model. The early defense of~\cite{papernot2016distillation} starts with an incorrect conjecture that ascribes adversarial example to high nonlinearity/overfitting, and develops defensive distillation to reduce the variations around input. The method is quickly subverted by~\cite{carlini2016defensive,carlini2017towards,athalye2018obfuscated} as argued in~\cite{fgsm} that the primary cause is due to local linearity of neural networks instead.

Hence, a large body of works focus on adversarial training~\cite{pgd,croce2020reliable,shafahi2019adversarial,wong2020fast} by solving a min-max saddle point problem. However, it is non-trivial to tackle the trade-off between robustness and accuracy~\cite{zhang2019theoretically}, which often leads to significant loss on clean image accuracy, with extensive training efforts. Applying adversarial training into the deep hashing domain suffers from even higher accuracy loss as we have experimented (see appendix). For image retrieval system, as long as the adversarial images are detected at the input, we can equivalently thwart the attacks without accuracy loss and training complexities.

Adversarial detections extract the artifacts left by the adversarial examples at different levels: raw pixels~\cite{gong2017adversarial,grosse2017statistical}, feature distributions~\cite{grosse2017statistical,li2017adversarial}, softmax distributions~\cite{hendrycks2016baseline} and frequency components~\cite{wang2020high}. By analyzing the contrastive distributions of the adversarial and natural images, a detector can be efficiently trained in a supervised or unsupervised manner. Another thread of works rely on the prediction inconsistency by exploiting denoise method and measuring the disagreement between the results~\cite{meng2017magnet,xu2017feature,wangxiaofeng}. All these works are based on softmax classification. In this work, we discover adversarial behaviors from the hamming space and propose a set of detection criteria including defending against the strongest white-box attackers.

\vspace{-0.05in}
\section{Conclusion}
In this paper, we propose an efficient, unsupervised detection of adversarial examples in deep hashing based image retrieval. We design three criteria to identify adversarial behaviors of both targeted and untargeted attacks in the hamming space and consider white-box attackers who are aware of the existence of the defense. The extensive evaluations demonstrate that the proposed detection can surpass previous defense techniques by a large margin and is also robust against white-box attacker by limiting its action space.

{\small
\bibliographystyle{ieee_fullname}
\bibliography{Bibliography-File}
}

\section{Appendix}
We provide additional experiments of adversarial training and white-box attacks in the Appendix.

\subsection{Adversarial Training}
As a proactive defense, adversarial training aims to solve a min-max optimization problem, that the inner optimization finds adversarial examples to maximize the loss function, whereas the outer optimization minimizes the overall loss. Thus, the trade-off between model accuracy and robustness becomes the key in adversarial training. In the following, we show that such gap becomes even larger in deep hashing and is non-trivial to handle with the conventional adversarial training method.

\textbf{Implementation Details.} Our implementation is based on Free Adversarial Training (FreeAT)~\cite{shafahi2019adversarial}. It parallelizes parameter update and adversarial generation, thus reducing the overhead from the inner optimization. During training, it replays the same minibatch data $m$ times to update the model parameters as well as generates adversarial examples for the next iteration. Small $m$ has little effect on robustness but large one would impact on the model accuracy. From~\cite{shafahi2019adversarial}, the best case of softmax classification (when $m=8$) has $0.46$ accuracy gain from the adversarial attacks (PGD and CW) and loses about $0.1$ accuracy on CIFAR-10. Next, we validate if similar value holds for deep hashing.

We adopt the FreeAT method in deep hashing using their configuration on CIFAR-10. We set $||\epsilon||_\infty=8$ , with four different repeat times $m$ from $2$ to $16$, to train four ResNet50 models. For each one of them, we generate PGD adversarial examples with various steps, denoted as PGD-8, PGD-40 and PGD-100 for $8$, $40$ and $100$ steps, respectively.

%Their perturbations $\epsilon$ are limited with $||\epsilon||_\infty=8$ as well, while their step sizes are all set to 1. %(8, 40, and 100 steps noted as PGD-8, PGD-40 and PGD-100, respectively)

%To adopt this method, we use a CIFAR-10 dataset for an example. $\epsilon$ is set to 8.0 and we set different repeat times ($m$) to explore the behaviors from different models.

\begin{table}[!ht]
\vspace{-0.08in}
\centering
\small
\hspace{-0.08in}\begin{tabular}{l | c | c c c @{}}
\toprule

%& \multicolumn{2}{c}{CIFAR-10} &  \multicolumn{2}{c}{ImageNet} &  \multicolumn{2}{c}{MS-COCO} &  \multicolumn{2}{c}{NUSWIDE} & \multicolumn{2}{c}{\emph{Avg. Gain}}  \\
%\cmidrule(lr){2-3} \cmidrule(lr){4-5} \cmidrule(lr){6-7} \cmidrule(lr){8-9} \cmidrule(lr){10-11}
%&Untgt &Tgt    &Untgt &Tgt    &Untgt &Tgt    &Untgt &Tgt     &Untgt &Tgt\\ %\hline
\multirow{2}{*}{Training} &  \multicolumn{4}{c}{Evaluated Against}\\
             &Benign &PGD-8& PGD-40 &PGD-100 \\
\midrule
No FreeAT      &0.854  & 0.028 & 0.030  & 0.033 \\% m=1 is 0.856 (Not a big difference). Exps on 0.856 model.
FreeAT $m=2$   &\textbf{0.889}  & 0.070 & 0.035  & 0.035 \\
FreeAT $m=4$   &0.838  & 0.222 & 0.070  & 0.055 \\
FreeAT $m=8$   &0.769  & \textbf{0.328} & \textbf{0.146}  & \textbf{0.130} \\
FreeAT $m=16$  &0.683  & 0.278 & 0.122  & 0.114 \\
 \hline
\end{tabular}
\caption{Clean and adversarial mAPs of implementing FreeAT~\cite{shafahi2019adversarial} for deep hashing on CIFAR-10}  \label{tab:adv_train}
%\vspace{-0.15in}
\end{table}

\textbf{Results.} The first row of Table~\ref{tab:adv_train} shows that without adversarial training, PGD attacks have successfully lowered the mAP to near zero. With FreeAT, the mAP improves under PGD-8, but still performs poorly with large attack steps such as PGD-40 and PGD-100. For the best case here when $m=8$, PGD attack with 100 steps still manages to lower the mAP to $0.130$. Not only having such low fidelity in defense, adversarial training also brings down the model accuracy on benign samples. With more $m$ steps involved, the mAP on benign samples decreases drastically from $0.854$ to $0.683$\footnote{The only exception is the model with $m=2$, which improves the performance on benign samples. It is inline with~\cite{xie2020adversarial}, where adversarial training is found to improve accuracy on clean images for some parameter settings.}, while no significant improvement of adversarial robustness is found. Compared to softmax classification, the gap is widening for deep hashing so we pursue adversarial detection in this paper.

\subsection{White-box Experiments}

We present additional results of the white-box attacks. Since the attacker need to design specific objectives regarding the detection methods, we choose MeanBlur~\cite{li2017adversarial} as a benchmark for comparison. The detection rate is True Positive Rate (TPR) with a certain threshold calculated by False Positive Rate (FPR) on benign samples. We use the PGD attack with $||\epsilon||_\infty=32$. For the best success rate and ability to bypass the detection, we choose $\lambda_1=0.0075$. In Table~\ref{tab:tpr_white}, we show the detection results on four datasets evaluated by TPR. Given FPR$=0.1$ and FPR$=0.2$, our method achieves an average of $0.7309$ and $0.8403$ white-box detection rates, whereas the detection rates for MeanBlur is close to zero. This further validates that our method is robust against the strongest white-box attackers. It is also interesting to see that both methods perform relatively better on single-label datasets (CIFAR-10 and ImageNet) than multi-label datasets (MS-COCO and NUSWIDE), which is different from the gray-box attacks. The reason behind that need to be further studied in the future.

\begin{table}[!ht]

\vspace{-0.08in}
\centering
\small
\begin{tabular}{l |c | c c c c@{}}
\toprule
%Detector &Loss\ref{eq:adapt_qloss} &Loss\ref{eq:adapt_denoiser2} &Loss\ref{eq:adapt_combine} &BIM (Untargeted) &FGSM (Untargeted) &C\&W  \\
%Detector & & CIFAR-10 & & & ImageNet &&& MS-COCO &&& NUSWIDE &

Detecotr & FPR& \multicolumn{1}{c}{CIFAR10} & \multicolumn{1}{c}{ImageNet} & \multicolumn{1}{c}{MS-COCO} & \multicolumn{1}{c}{NUSWIDE} \\
\midrule
%\multirow{2}{7mm}{\begin{sideways}\parbox{10mm}{MeanBlur}\end{sideways}}
\multirow{2}{*}{MeanBlur~\cite{li2017adversarial}}&0.1 &0.0340 &0.0218 &0.0044 & 0.0057  \\
    &0.2 &0.0480 &0.0250 &0.0076 &0.0076\\
\hline
%\multirow{2}{7mm}{\begin{sideways}\parbox{10mm}{Ours}\end{sideways}}
\multirow{2}{*}{Our Method}&0.1 &0.9560 &0.8824 &0.4648 & 0.6204  \\
    &0.2 &0.9790 &0.9180 &0.6742 & 0.7900 \\
\hline
\end{tabular}
\caption{Detection rate (TPR) of the adversarial examples against white-box attacks (PGD, $\epsilon=32$, $\lambda_1=0.0075$)}
\vspace{-0.1in}
\label{tab:tpr_white}
\end{table}

\end{document}